\documentclass[letterpaper, 10 pt, journal, twoside]{IEEEtran}
\usepackage{amsmath}
\usepackage{amssymb}
\usepackage{graphicx}
\usepackage{url}
\usepackage{cite}

\usepackage{booktabs}
\usepackage{array}
\usepackage{multirow}
\usepackage{colortbl}
\usepackage[table]{xcolor}

\begin{document}
%
\title{Learning Dynamic Pick-and-Place for a Legged Manipulator}
%
%
%

\author{Moonkyu Jung$^{1}$, Jiseong Lee $^{1}$, Zhengmao He$^{2}$, Donghoon Youm$^{1}$, Juhyeok Mun$^{1}$, HyeongJun Kim$^{1}$, Hyunsik Oh$^{1}$, Donghyuk Choi$^{1}$, Jungwoo Hur$^{1}$, Jie Song$^{*,2}$ and Jemin Hwangbo$^{*,1}$
\thanks{Manuscript received: December 15, 2025; Revised March 7, 2026; Accepted April 12, 2026.}
\thanks{This paper was recommended for publication by Editor Wei Pan upon evaluation of the Associate Editor and Reviewers' comments.
This research was financially supported by the Institute of Civil Military Technology Cooperation funded by the Defense
Acquisition Program A dministration and Ministry of Trade, Industry and Energy of Korean government under Grant UM22207RD2. \textit{($^{*}$Corresponding author: Jie Song and Jemin Hwangbo)}} 
\thanks{$^{1}$Moonkyu Jung, Jiseong Lee, Donghoon Youm, Juhyeok Mun, HyeongJun Kim, Hyunsik Oh, Donghyuk Choi,  Jungwoo Hur and Jemin Hwangbo are with Robotics and Artificial Intelligence Lab, KAIST, Daejeon, South Korea {\tt\footnotesize moonk127, ssuni1002016, ydh0725, munjuhyeok, kaist0914, ohsik1008, cay879, jungwoohur, jhwangbo@kaist.ac.kr}}%
\thanks{$^{2}$Zhengmao He and Jie Song are with the Thrust of Robotics and Autonomous Systems, The Hong Kong University of Science and Technology (Guangzhou), Guangzhou, China {\tt\footnotesize zhmaohe@gmail.com, jsongroas@hkust-gz.edu.cn}}%
\thanks{Digital Object Identifier (DOI): see top of this page.}
}
%
%

\markboth{IEEE Robotics and Automation Letters. Preprint Version. Accepted April, 2026}
{Jung \MakeLowercase{\textit{et al.}}: Learning Dynamic Pick-and-Place for a Legged Manipulator} 

%



\maketitle

\begin{abstract}
Legged manipulators extend robotic capabilities beyond static manipulation by integrating agile locomotion with versatile arm control. However, achieving precise manipulation while maintaining coordinated locomotion remains a major challenge. This work presents a hierarchical reinforcement learning framework for dynamic pick-and-place tasks using a quadruped equipped with a 6-DOF robotic arm. The framework incorporates an explicit mass estimation module enabling adaptive whole-body control for objects with varying weights. In simulation, the system achieves an 86.05\% success rate with payloads up to 2.3 kg. The approach is further validated through real-world experiments across six representative scenarios with controlled variations in object physical properties (size and mass) and task heights. Specifically, within a wide vertical workspace ranging from ground level to 1.1~m-high tabletops, the system demonstrates an average success rate of 73.3\% for payloads up to 1.3 kg, with an average execution time of 4.06 s. Unlike prior works that handle lightweight objects and execute pick-and-place motions with slow, piecewise motions, the proposed framework exploits concurrent locomotion and manipulation for dynamic, continuous execution. These results demonstrate the potential of quadrupedal mobile manipulators for adaptive, whole-body pick-and-place with heavier payloads and extended workspaces.
\end{abstract}


%
\IEEEpeerreviewmaketitle

\vspace{-0.40cm}

\section{Introduction}
%
%
%
%
\IEEEPARstart{L}{egged} manipulators—robots integrating a quadrupedal base with a robotic arm—have the potential to substantially extend pick-and-place capabilities beyond the structured, planar environments that constrain traditional fixed-base and wheeled manipulators. Their quadrupedal bases provide high degrees of freedom and adaptive mobility, enabling a wide range of dynamic behaviors from stable walking to agile parkour-like motions \cite{hoeller2024anymal, kim2025high}.
Building on this potential, our goal is to design a controller that enables fast and accurate pick-and-place operations by fully exploiting the quadruped’s locomotion and whole-body posture during object manipulation. However, extending such dynamic motions to manipulation tasks poses several technical challenges. Oscillations generated during locomotion are amplified at the distal end-effector, complicating precise manipulation. In addition, arm motions induce wrench disturbances at the base, hindering the maintenance of stable posture.

\begin{figure}[t!]
    \centering
    \includegraphics[width=\columnwidth, trim=0cm 0cm 0cm 0cm, clip]{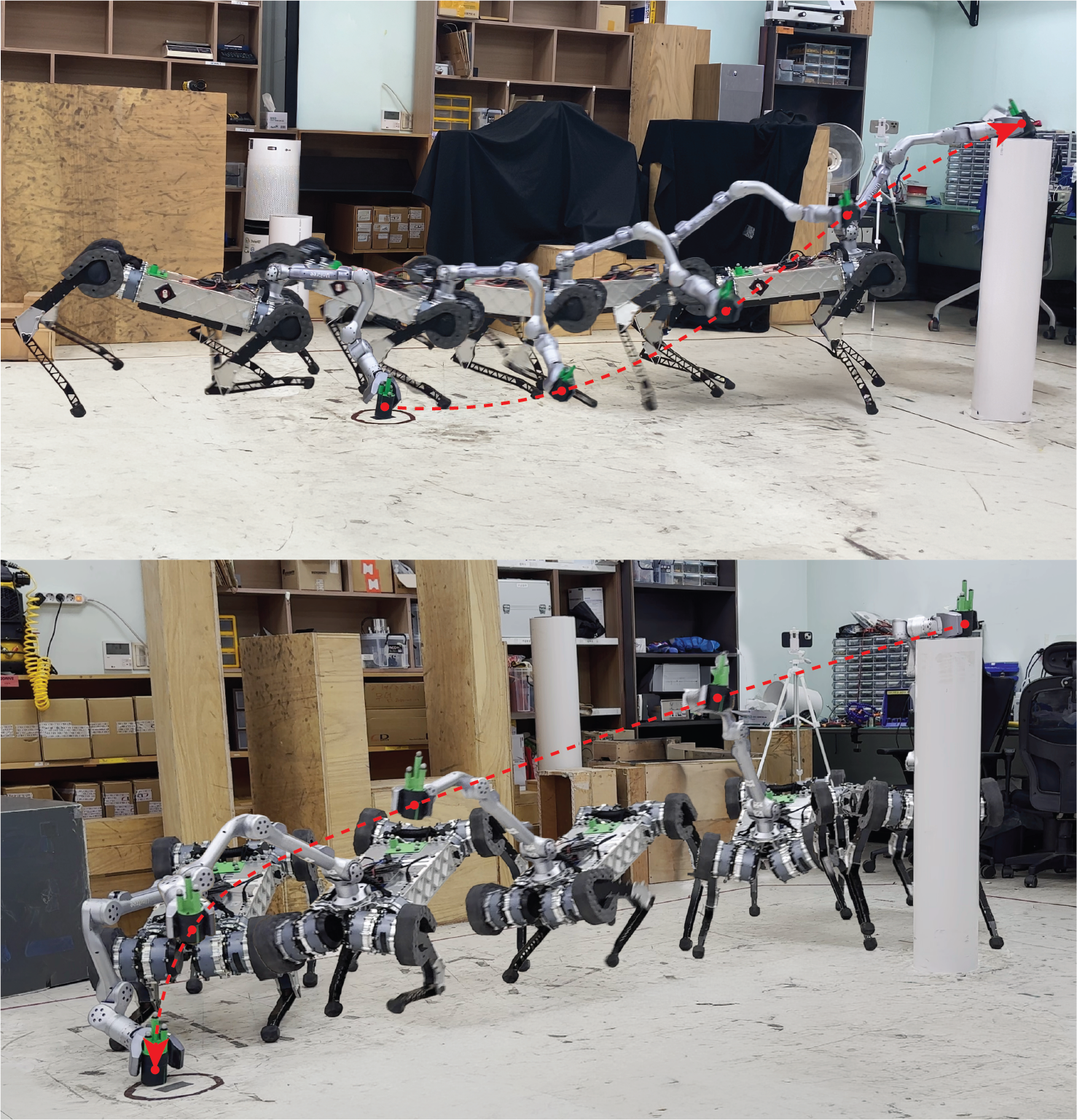}
    \vspace{-0.50cm}
    \caption{\textbf{Dynamic Pick-and-Place of Legged Manipulator}: Sequential snapshots capturing real-world deployments using the proposed framework.
    The \textbf{top} sequence shows transferring an object from the floor to a 1.1~m-high tabletop,
    and the \textbf{bottom} sequence shows placing an object from the tabletop back onto the floor.
    These demonstrate the framework’s capability to coordinate agile locomotion and precise manipulation,
    enabling rapid, whole-body pick-and-place across large height differences.}
    \label{fig: intro}
    \vspace{-0.60cm}
\end{figure}

To address these issues, earlier research has primarily relied on model-based optimal control methods, such as Model Predictive Control (MPC), for quadrupedal loco-manipulation. These approaches utilize accurate dynamic models of both the robot and its environment to achieve precise motion planning and control \cite{sleiman2021MPC, chiU2022uMPC}. However, as task complexity increases, the computational burden of online optimization grows rapidly, often impeding real-time execution. Moreover, these methods depend on accurate modeling of object and contact dynamics, which is difficult to achieve in unstructured environments, thereby restricting their applicability to real-world scenarios.

To overcome these limitations, reinforcement learning (RL) has recently emerged as a promising alternative for quadrupedal loco-manipulation \cite{fu2022deep, liu2024visual, portela2024end, portela2024force, zhang2024gamma, yokoyama2024asc}. Model-free RL leverages large-scale simulation to acquire control policies that generalize across diverse environments without explicit system modeling. Despite these advances, existing RL-based approaches remain largely confined to tasks involving the manipulation of light objects and are typically designed by training separate controllers for each stage of the pick-and-place process \cite{zhang2024gamma, yokoyama2024asc}. As a result, they fail to fully exploit the integrated capabilities of quadrupedal locomotion and arm manipulation, leading to quasi-static and inefficient task execution.

In this study, we propose a learning-based controller for dynamic pick-and-place with a quadrupedal mobile manipulator that addresses these limitations. Our approach adopts a hierarchical structure: a low-level controller is trained to provide robust locomotion and whole-body stability under diverse arm movements, while a high-level controller is trained to perform task execution and arm control on top of this stable locomotion base. This decomposition of the action space enables efficient and stable policy learning in high-dimensional systems. Moreover, by training the entire pick-and-place process as a single integrated episode—including grasping, transport, placement, and retreat—the framework ensures seamless coordination across task stages. To further enhance generalization, we employ a success-rate-driven curriculum that progressively increases task complexity, allowing the robot to manipulate objects placed on tabletop surfaces exceeding twice its base height. Finally, we introduce an adaptation module that estimates object mass through physical interaction during execution, enabling robust performance across a broad range of object weights—from light to relatively heavy loads.

In summary, the main contributions of this work are as follows:
\begin{itemize}
 \item We propose a single unified high-level policy for long-horizon pick-and-place tasks by leveraging a robust low-level whole-body controller, enabling fast and seamless coordination across task stages.
 \item We incorporate a success-rate-driven curriculum and an online mass-adaptation module, allowing dynamic manipulation of relatively heavy objects and operation over a wide vertical range.
 \item We validate the proposed framework in both simulation and real-world experiments.
\end{itemize}
To the best of our knowledge, this work presents the first demonstration of rapid and continuous pick-and-place operations in legged manipulation while maintaining high base velocity during manipulation.

\section{RELATED WORKS}  
\begin{figure*}[!t]
    \centering
    \vspace{0.2cm}
    \includegraphics[width=\textwidth, trim=9.9cm 7.9cm 15.7cm 3.7cm, clip]{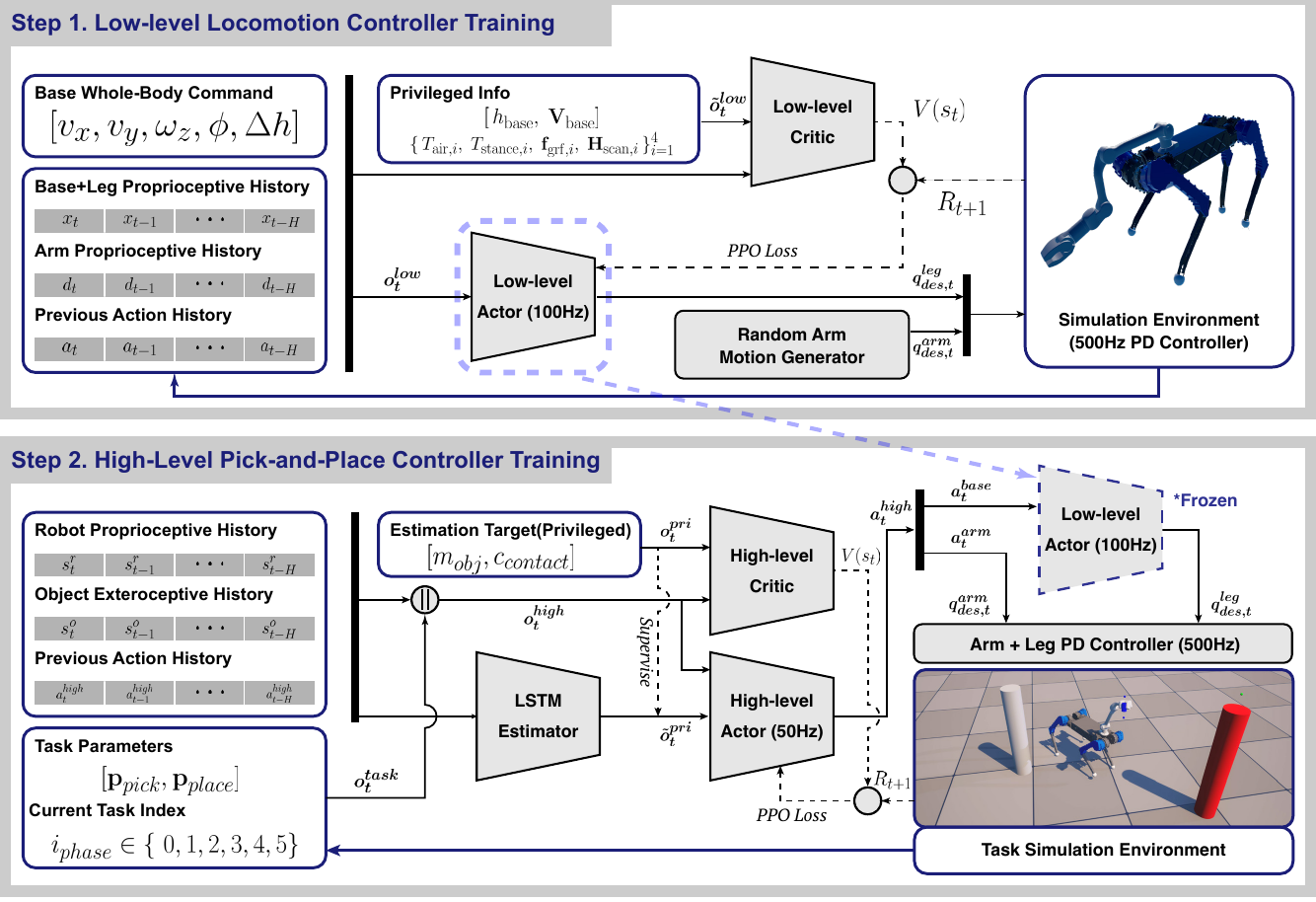}
    \vspace{-0.7cm}
  \caption{\textbf{Hierarchical training framework:} Our framework consists of two          training stages. In \textbf{Step~1}, a low-level locomotion controller is trained via   reinforcement learning to generate robust whole-body stabilization under random arm     disturbances. 
    In \textbf{Step~2}, the low-level actor is frozen, and a high-level pick-and-place controller is trained jointly with an LSTM estimator that predicts object mass and contact state from proprioceptive and exteroceptive histories. 
    The high-level action is defined as 
    $a^{high} = [a^{base}, a^{arm}]$, 
    where $a^{base}$ represents the locomotion command required for the current task and $a^{arm}$ denotes the desired arm motion. 
    Specifically, the base command consists of the target base velocity, pitch angle, and height: 
    $a^{base} = [v_x, v_y, \omega_z, \phi, \Delta h]$. 
    The low-level controller converts $a^{base}$ into desired leg joint positions $q^{leg}_{des}$ for 12 actuated joints, ensuring stable locomotion, 
    while $a^{arm}$ produces desired arm joint positions and gripper commands $q^{arm}_{des}$. 
    Joint targets are converted into torque commands using joint-space PD control.}
    \label{fig:overall}
    \vspace{-0.7cm}
\end{figure*}

\subsection{Whole-Body Controllers for Legged Manipulators}
Whole-body control (WBC) provides a framework for coordinating locomotion and manipulation in legged manipulators. Early works \cite{sleiman2021MPC, chiU2022uMPC} leveraged Model Predictive Control (MPC) for tasks like door opening and object throwing, but suffered from high computational costs and dependence on accurate models.

To improve terrain adaptability, Ma et al.~\cite{ma2022combine} combined RL-based locomotion with MPC planning, though it still required accurate models and lacked unified coordination between the arm and base.
In contrast, Fu et al.~\cite{fu2022deep} proposed a fully unified RL policy for simultaneous base and end-effector control. Expanding on this, Portela et al. achieved high-accuracy 6D pose tracking and force control on complex terrains using keypoint representations and dense-sparse reward structures in their recent works \cite{portela2024end, portela2024force}.

\subsection{Learning-Based Task Execution for Legged Manipulators}
With the advancement of learning-based control, RL-based frameworks have been applied to specialized tasks such as door opening, badminton, and ball throwing \cite{zhang2024learning, ma2025learning, fey2025bridgingsimtorealgapathletic}.

For grasping-based manipulation, Liu et al.~\cite{liu2024visual} and Zhang et al.~\cite{zhang2024gamma} developed high-level policies for robust grasping in diverse environments using point-cloud features or graspability estimation. However, these approaches focused only on grasp execution and did not address subsequent transport or placement with whole-body coordination.

For full pick-and-place, Yokoyama et al.~\cite{yokoyama2024asc} decomposed the task into modular visual-motor skills, but their reliance on the robot's internal controller resulted in slow execution and prevented unified locomotion-manipulation coordination. Similarly, Zhang et al.~\cite{zhang2025learning3} distilled multi-stage teacher policies into a single student policy for unified control. Although they demonstrated vision-based autonomous pick-and-place, the execution remained quasi-static, averaging 43.8\,s per task and being largely limited to simple floor-level drop-off tasks.

In contrast, our work focuses on enhancing the dynamic capability of the legged manipulator, placing greater emphasis on whole-body coordination than on perceptual sophistication. By exploiting concurrent locomotion and manipulation, our framework reduces the average real-world execution time to 4.06\,s, achieving a 10-fold speedup over prior quasi-static approaches. It also extends the workspace to a 1.1\,m-high tabletop—more than twice the robot's base height—and enables precise upright placement of payloads up to 1.3\,kg.

\section{METHOD}
\subsection{Overview}
\subsubsection{\textbf{Dynamic Pick-and-Place}}

We define the dynamic pick-and-place task as a continuous, whole-body motion problem requiring a legged manipulator to grasp, transport, and place an object between two spatially distinct locations.

The environment comprises two cylindrical tables (10 cm radius): a \textit{pick table} and a \textit{place table}, visualized in Fig.~\ref{fig:overall} in white and red, respectively. The target object, either a cylinder or a cuboid, is placed upright on the pick table. To ensure robustness, the task setup is randomized at the start of each episode: the pick table is spawned at a random location relative to the robot, and the place table is spawned at a random location and height relative to the pick table. The control objective is to grasp the object, transport it, and place it upright near the center of the place table. The entire sequence must be executed as a single, continuous motion without pauses for regrasping or corrective adjustments. After successful placement, the robot must retract its arm to a neutral posture without disturbing the object. This task setup introduces two primary challenges that distinguish it from conventional manipulation studies:

\begin{enumerate}
\renewcommand\theenumi{\roman{enumi}}
  \renewcommand\labelenumi{(\theenumi)}
    \item Dynamic Grasping with Locomotion: Unlike quasi-static approaches that minimize impact by approaching objects slowly, our task requires grasping while the robot's base is in motion. This demands precise coordination between agile locomotion and end-effector control—specifically, maintaining whole-body balance while minimizing the relative velocity between the end-effector and the object at the moment of contact.
    \item Adaptive Whole-Body Posture Control: Unlike single-plane tabletop tasks, our setup involves transferring objects between surfaces with significant height differences. This necessitates robust and adaptive whole-body posture control throughout the entire motion, dynamically coordinating the legs and arm to reach disparate locations.
\end{enumerate}

\subsubsection{\textbf{Hierarchical Policy Structures}}
Directly training a single, monolithic policy to solve this task poses significant learning challenges. First, the coupling between the arm and base (related to challenges (i) and (ii)) is problematic during early training; unstable base motion causes the arm policy to focus on avoiding instability rather than achieving the manipulation goal, often resulting in conservative postures. Second, a single policy must explore a vast 19-DoF action space (12 for the legs, 7 for the arm/gripper), which greatly hinders learning efficiency.

To address these issues, we adopt a hierarchical control structure as shown in Fig.~\ref{fig:overall}. A \textit{low-level controller} is trained to stabilize base motion and regulate whole-body posture under diverse arm disturbances, providing a reliable foundation for higher-level learning. Building upon this, a \textit{high-level controller} is trained to generate task-level commands that directly control both locomotion and manipulation. This separation allows the high-level controller to focus on efficient task execution while relying on the low-level controller for stable base dynamics.

\subsection{Low-level Controller Learning}
\subsubsection{\textbf{Whole-Body Locomotion Controller}}

The low-level controller generates desired leg joint positions for whole-body locomotion and body stabilization. 
It receives the base velocity command $[v_{x}, v_{y}, \omega_{z}]$ and the body-control command $[\Delta h, \phi]$, and outputs desired joint positions for the 12 actuated leg joints ${q^{leg}_{des,t}}$. The policy receives the low-level observation $o^{\text{low}} = [\, x, d, a \,]$, where $x$ denotes the proprioceptive state of the base and legs, $d$ represents the arm’s proprioceptive state, and $a$ denotes the previous action.
The overall locomotion framework follows the reward design introduced by Ji et al. \cite{ji2022concurrent}, while the body-control command specifically regulates the base height and pitch angle based on the formulations of Margolis et al. \cite{margolis2022walktheseways}. To enhance training efficiency and stability, we employ an asymmetric actor–critic framework \cite{pinto2018asymmetric}, providing the critic with privileged observations $[h_{base}, {V}_{base},{T}_{air}, {T}_{stance}, {f}_{grf}, H_{scan}]$, which represent, in order, the base height, base linear velocity, air and stance times of each leg, ground reaction forces, and height-scan values around each foot.

Prior works \cite{fu2022deep, liu2024visual} achieved low-level whole-body control by generating both arm and base motions from the end-effector target in a ground-fixed frame, resulting in a single, uniquely determined base posture for each target.
In contrast, our framework treats the base posture as a controllable variable, allowing the high-level controller to flexibly search for optimal base and arm configurations that satisfy the task objective.

\subsubsection{\textbf{Random Arm Motion Generator}}
The random arm-motion generator applies dynamic disturbances to the robot base in the form of wrenches at the arm-base interface.
Rather than explicitly estimating these external forces\cite{ma2022combine}, the proposed controller implicitly learns to compensate for them from arm state histories. To generate a wide range of generalized arm motions, each joint is assigned an independent target position within its feasible range, and the trajectory duration is randomly sampled per joint to induce asynchronous motion patterns. This contrasts with previous approaches \cite{fu2022deep, liu2024visual}, which assumed constant end-effector velocity and generated motions through linear interpolation in the end-effector space.

\newcommand{\dq}[1]{\Delta q_{#1}}
\newcommand{\Ti}[1]{T_{#1}}
\newcommand{\ai}[1]{a_{#1}}

At the beginning of each episode, the initial joint position \(q_{\text{init}, i}\), target position \(q_{\text{target}, i}\), 
and trajectory duration \(T_i\) are randomly sampled within each joint’s feasible range. 
The joint displacement is defined as \(\dq{i}=q_{\text{target}, i}-q_{\text{init}, i}\). 
Each joint \(i \in \{0, 1, 2, 3, 4, 5\}\) independently selects one of two motion modes: \textit{constant-velocity motion} or \textit{symmetric constant-acceleration motion}.

Since the trajectory duration $T_i$ is randomly sampled and often shorter than the episode length, 
certain joints remain stationary while others continue moving, leading to diverse and asynchronous arm motion patterns. 
These randomized trajectories encourage the controller to handle a wide range of dynamic arm–base interactions 
and enhance generalization in joint-space learning. 
Additionally, a random payload of up to 2~kg is attached to the end-effector during training 
to simulate various object weights encountered in manipulation tasks.

\subsection{High-level Controller Learning}
The high-level controller generates high-level actions $\boldsymbol{a}^{\text{high}}_t$ for performing the pick-and-place task. 
The action consists of the base command $a^{\text{base}}_t = [v_{x}, v_{y}, \omega_{z}, \Delta h, \phi]$, and the arm command $a^{\text{arm}}_t$. 
To ensure stability, $a^{\text{base}}_t$ is scaled via $\tanh(\cdot)$ and $\sigma(\cdot)$ activations to stay within the low-level controller's operative limits: $ -1 \le v_x \le 2$ m/s, $|v_y| \le 1$ m/s, $|\omega_z| \le 1$ rad/s, $|\phi| \le 0.28$ rad, and $-0.2 \le \Delta h \le 0$ m.
Regarding the arm, the command is defined as
$a^{\text{arm}} = [\, q^{\text{arm},0}_{\text{des}}, \ldots, q^{\text{arm},5}_{\text{des}}, a^{\text{gripper}} \,]$, 
where $q^{\text{arm},i}_{\text{des}}$ ($i = 0, 1, \ldots, 5$) denote the desired joint angles of the 6-DOF arm, and 
$a^{\text{gripper}}$ is a binary command indicating the gripper state (open or closed).

To generate these actions, the policy receives an observation 
$o^{\text{high}} = [\, s^r, s^o, a^{\text{high}}, p_{\text{pick}}, p_{\text{place}}, i_{\text{phase}} \,]$. 
Here, $s^r$ denotes the robot’s proprioceptive state, and $s^o$ represents the  object’s state. 
The previous action $a^{\text{high}}$ provides short-term behavioral context, while 
$p_{\text{pick}}$ and $p_{\text{place}}$ specify the object’s initial location and the target table position, both expressed in the robot's coordinate frame. 
The phase index $i_{\text{phase}}$ indicates the current stage of the task and helps the policy manage long-horizon behaviors. 
Among these, $[\, s^r, s^o, a^{\text{high}} \,]$ are additionally provided as historical sequences through a history buffer.

The object state $s^o_t$ is encoded using a keypoint-based formulation inspired by prior work \cite{portela2024end, jeon2023whole}, which employs richer keypoint sets for general 6D object representation. In our setting, however, the object is assumed to remain upright, allowing a compact three-keypoint parameterization that specifies the object's center and its top and bottom positions along the vertical axis. This representation does not capture the full $SE(3)$ pose (e.g., yaw) but provides a stable and task-appropriate alternative to explicit $SE(3)$ representation.
The state also includes the object’s size and a one-hot vector specifying its shape (cuboid or cylinder).

Finally, the phase index $i_{\text{phase}}$ provides coarse task-level progression cues and is defined as: 
(0) initial state, 
(1) end-effector above the pick table, 
(2) end-effector above the place table, 
(3) object placed on the place table, 
(4) object released, and 
(5) arm returned to the initial posture. 
While similar stage indices have been used in prior work \cite{zhang2025learning3}, we exclude explicit grasping conditions for phase transitions. This decision is motivated by our empirical observations; omitting strict grasping criteria yielded a higher overall success rate compared to relying on heuristic constraints.

\subsubsection{\textbf{Online Estimation Module for Adaptive Control}} While many previous studies focused on quasi-static pick-and-place with light payloads, our scenario requires manipulating objects near the arm’s 2.3 kg payload limit under tight timing constraints. This setup amplifies the influence of object mass on control performance. However, most robotic systems lack sensors capable of directly measuring object mass during runtime.

Extending the idea of inferring latent physical properties from state histories explored in prior works \cite{ji2022concurrent, fu2022deep, jeon2023whole} to prehensile manipulation, we define the object mass $m_{\text{obj}}$ and the gripper–object contact state $C_{\text{contact}}$ as privileged observations $\boldsymbol{o^{\text{pri}}}$. We then design an LSTM-based estimator to infer these parameters in real time as the robot executes the pick-and-place motion.

The estimator predicts $\boldsymbol{\tilde{o}^{\text{pri}}}$ from the robot–object interaction data $[\, s^{r}, s^{o}, a^{\text{high}} \,]$. These predictions are supervised during training using ground-truth privileged values $\boldsymbol{o}^{\text{pri}}$ obtained from the simulation, as illustrated in Step~2 of Fig.~\ref{fig:overall}. The resulting estimates are then incorporated into the actor's observation along with $o^{\text{high}}$.

\subsubsection{\textbf{Success-rate-driven Curriculum Learning}}

\begin{table}[!t]
\vspace{0.2cm}
\centering
\caption{Task Variables}
\vspace{-0.2cm}
{\renewcommand{\arraystretch}{1.1}
\begin{tabular}{|cc|c|}
\hline
\multicolumn{2}{|c|}{\textbf{Parameters}}                                    & \textbf{Training Range} \\ \hline
\multicolumn{1}{|c|}{\multirow{2}{*}{Pick location}}    & Distance        & $U~(0.9, ~3.0)~m$             \\ \cline{2-3} 
\multicolumn{1}{|c|}{}                                  & Height          & $U(0.0,~1.3)~m$          \\ \hline
\multicolumn{1}{|c|}{\multirow{2}{*}{Place location}}   & Distance        & $U(0.5,~3.0)~m$             \\ \cline{2-3} 
\multicolumn{1}{|c|}{}                                  & Height          & $U(0.0,~1.3)~m$           \\ \hline
\multicolumn{2}{|c|}{Object yaw}                                        & $U(0,~\pi)~rad$              \\ \hline
\multicolumn{2}{|c|}{Object type}                                            & ${[}Box,~Cylinder{]}$    \\ \hline
\multicolumn{1}{|c|}{\multirow{3}{*}{Object dimension}} & Box width & $U(3,~5)~cm$                  \\ \cline{2-3} 
\multicolumn{1}{|c|}{}                                  & Cylinder diameter         & $U(4,~7)~cm$                \\ \cline{2-3} 
\multicolumn{1}{|c|}{}                                  & Height         & $U(6,~10)~cm$                \\ \hline
\multicolumn{2}{|c|}{Object mass}                                        & $U(0.2~,2.3)~kg$              \\ \hline

\end{tabular}
}
\label{table:env}
\vspace{-0.4cm}
\end{table}

During high-level policy training, each episode begins by sampling the configurations for the pick and place tables, along with the object’s mass and size (Table~\ref{table:env}). All random variables follow uniform distributions, while the table radius remains fixed at 10~cm.

To improve training stability, we employ a success-rate-driven curriculum that gradually increases task difficulty. Three curriculum variables—denoted as $L_{\text{pick}}$, $L_{\text{place}}$, and $L_{\text{release}}$—are initialized near zero and increment toward 1.0 based on their respective subgoal success rates. These levels directly modulate the sampling ranges of the task parameters.

Task difficulty is controlled through (i) the robot-to-pick-table distance, (ii) the inter-table distance, (iii) the object mass, and (iv) the success criteria for the place and release subgoals. Training begins with all levels set to 0.10, corresponding to short-range, light-object tasks with relaxed criteria. Every 50 iterations, each level increases by 0.01 if the corresponding success rate exceeds a predefined threshold.

To prevent unbalanced progression (e.g., mastering picking while rarely attempting placing), we introduce a synchronization constraint: $L_{\text{pick}}$ and $L_{\text{place}}$ are updated only if they remain within a specified margin of each other. In our implementation, the success thresholds for the pick, place, and release subgoals are 0.90, 0.85, and 0.85, respectively, with a regulation margin of 0.015.

\begin{table}[t!]
\centering
\vspace{0.2cm}
\scriptsize
\caption{Reward Terms}
\vspace{-0.2cm}
\renewcommand{\arraystretch}{1.5}

\resizebox{\columnwidth}{!}{
\begin{tabular}{|c|c|p{0.7 \columnwidth}|}
\hline
\textbf{Stage (Sparse/Dense)} & \textbf{Reward}      & \textbf{Expression} \\ \hline
\multirow{2}{*}{Pre-grasping (D)} 
    & \textit{EE\_to\_Obj}       & $k_1 \cdot e^{-25 \cdot \|p_{\text{obj}} - p_{\text{ee}}\|^2} + k_2 \cdot e^{-1 \cdot \|p_{\text{obj}} - p_{\text{ee}}\|^2}$ \\ \cline{2-3}
    & \textit{EE\_Obj\_Contact}  & $k_3 \cdot [\text{ee\_obj\_contact}]$ \\ \hline

Grasping (S) & \textit{Grasping Success} & $k_4 \cdot ([\text{object in gripper}] + [\text{grasping success}])$ \\ \hline

\multirow{3}{*}{Carrying (D)} 
    & \textit{Base\_Heading}      & $k_5 \cdot \mathbf{R}_{\text{base}}^{(x) T} \cdot \mathbf{d}_{\text{base} \to \text{place} }$ \\ \cline{2-3}
    & \textit{Obj\_to\_Place}    & $k_6 \cdot e^{-5 \cdot \|p^{key}_{\text{obj}} - p^{key}_{\text{place}}\|^2} + k_7 \cdot e^{-25 \cdot \|p^{key}_{\text{obj}} - p^{key}_{\text{place}}\|^2}$ \\ \cline{2-3}
    & \textit{Base\_to\_Place}   & $k_8 \cdot e^{-0.5 \cdot max\{\|p_{\text{base}} - p_{\text{place}}\|, 0.3\}}$ \\ \hline

\multirow{2}{*}{Placement (S)} 
    & \textit{Place Success}     & $k_9 \cdot [\text{placement success}]$ \\ \cline{2-3}
    & \textit{Gripper Release}   & $k_{10} \cdot [\text{gripper open}]$ \\ \hline

\multirow{2}{*}{Retreating (D)} 
    & \textit{Base Retreat}      & $k_{11} \cdot \min(1, \|p_{\text{base}} - p_{\text{place}}\|^2)$ \\ \cline{2-3}
    & \textit{EE Retreat}        & $k_{12} \cdot (1 - e^{-5 \cdot \min(1, \|p_{\text{ee}} - p_{\text{object}}\|)})$ \\ \hline

Finishing (S) & \textit{Complete}         & $k_{13} \cdot ([\text{release success}] + 2 \cdot [\text{retreat success}])$ \\ \hline

\multicolumn{3}{|c|}{\textbf{Penalty (Negative) Reward}} \\ \hline
\textit{Arm Penalty} & \multicolumn{2}{p{0.8 \columnwidth}|}{
    $k_{14} \cdot \sum \| \dot{q}_{i} \| $ + $k_{15} \cdot \| a^{arm}_{t}-a^{arm}_{t-1} \| $ \newline 
    + $k_{16} \cdot \sum \| \tau_{i} \|$ + $k_{17} \cdot \Sigma \|q_{i}-q_{i,nominal} \| $ \newline 
    + $
     k_{18} \cdot \sum \mathbb{I}\left( |q_i - q_{i,\text{min}}| \leq 0.05 \cdot q_{i,\text{min}} \, \text{or} \, |q_i - q_{i,\text{max}}| \leq 0.05 \cdot q_{i,\text{max}} \right)$
} \\ \hline
\textit{Base Penalty} & \multicolumn{2}{p{0.8 \columnwidth}|}{
    $k_{19} \cdot (V^2_z + 0.02\cdot |\omega_x| +0.02\cdot |\omega_y|)$ + $k_{20} \cdot \| c_{t}-c_{t-1} \|^2
    $\newline 
    + $k_{21} \cdot \| c_{t}-2c_{t-1}+c_{t-2} \|^2$+ $k_{22} \cdot (1 + \|b_{t}\|^2)  \cdot \|c_{t}\|$\newline 
    + $k_{23} \cdot \| b_{t}-b_{t-1} \|^2
    $ + $k_{24} \cdot \| b_{t}-2b_{t-1}+b_{t-2} \|^2$ + $k_{25} \cdot \|b_{t}\|^2$
} \\ \hline
\textit{Manipulation Penalty} & \multicolumn{2}{p{0.8 \columnwidth}|}{
     $k_{26} \cdot \| F_{g-t} \|^2$ + $k_{27} \cdot e^{-\| \mathbf{p}_{\text{obj}} - \mathbf{p}_{\text{place}} \|}  \cdot ( \mathbf{R}_{\text{obj}}^{(z)} - 1 )^2 $
} \\ \hline
\end{tabular}}
\vspace{-0.6cm}
\label{table:reward}
\end{table}

\begin{table}[t]
\centering
\vspace{0.2cm}
\scriptsize
\caption{Notation for Reward Formulation}
\vspace{-0.2cm}
\renewcommand{\arraystretch}{1.3} 
\resizebox{\columnwidth}{!}{%
\begin{tabular}{|l|l||l|l|}
\hline
\textbf{Symbol} & \textbf{Description} & \textbf{Symbol} & \textbf{Description} \\ \hline
$p_{\{\cdot\}}$ & Pos. of the corresponding entity & 
$q_i, \dot{q}_i$ & Arm joint pos. \& vel. \\ \hline

$p^{key}_{\{\cdot\}}$ & Set of keypoint positions & 
$a^{arm}$ & Arm action \\ \hline

$[\,\cdot\,]$ & Boolean indicator of state & 
$\tau_i$ & Arm joint torque \\ \hline

$\mathbf{R}_{\text{base}}^{(x)}$ & Base $x$-axis heading vector & 
$q_{i,\text{min/max}}$ & Arm joint position limits \\ \hline

$\mathbf{d}_{\text{base} \to \text{place}}$ & Dir. vector from base to place & 
$c_t$ & Base velocity command \\ \hline

$V_z$ & Base vertical velocity & 
$b_t$ & Base motion cmd. (pitch, height) \\ \hline

$\omega_{x}, ~\omega_{y}$ & Base roll, pitch velocity & 
$F_{\text{g-t}}$ & Gripper-table contact force \\ \hline

$\mathbf{R}_{\text{obj}}^{(z)}$ & Object $z$-axis vector & 
- & - \\ \hline
\end{tabular}
}
\label{table:nomenclature}
\vspace{-0.6cm}
\end{table}

\subsubsection{\textbf{Reward Design}}

Training a high-level pick-and-place policy involves a long-horizon decision-making process in which multiple subgoals must be achieved sequentially. Without careful reward shaping, value estimation can become unstable across task phases, often causing the policy to converge to local optima in early stages and fail to complete the task. To mitigate this, we design a structured reward function that decomposes the overall task into six stages—pre-grasping, grasping, carrying, placement, retreating, and finishing. This structure provides stage-specific guidance while preserving clear success signals.

In each stage, we combine sparse and dense rewards to balance explicit success criteria with smooth shaping terms, and we assign reward terms only to the subgoal relevant to the current stage. Reward components associated with earlier stages are simply no longer active once their corresponding subgoals are achieved, ensuring that each stage is guided solely by its intended objectives.

Table~\ref{table:reward} summarizes the reward components for each stage. Dense shaping terms, such as \textit{EE\_to\_Obj}, \textit{Obj\_to\_Place}, and \textit{Base\_to\_Place}, guide the agent during approach and transportation phases. Conversely, sparse rewards—including \textit{Grasping Success}, \textit{Place Success}, and \textit{Complete}—serve as discrete signals marking successful transitions between stages. Additionally, penalty terms regularize joint velocities, torques, base motions, and excessive contact forces to ensure stable operation throughout the task. The mathematical notation for the reward formulation is detailed in Table~\ref{table:nomenclature}.

\section{EXPERIMENTS AND RESULTS}

We evaluated the robustness of the low-level controller and the task performance of the high-level controller in both simulation and real-world environments. Through quantitative and qualitative comparisons against several baseline methods, we demonstrate the validity and effectiveness of the proposed system.

Experiments were conducted using an in-house quadruped platform integrated with a Unitree Z1 manipulator (6-DoF arm with a 1-DoF gripper), offering an effective payload of about 2.3 kg after considering the gripper weight.

Regarding the network architectures, the low-level controller is implemented as a MLP with hidden dimensions of [256, 128] using LeakyReLU activations. The high-level controller employs a MLP with [512, 512, 128] hidden units and LeakyReLU activations. Additionally, the estimator utilizes a two-layer LSTM with a hidden dimension of 24. Both policies are trained using PPO\cite{schulman2017proximal}.

All training was conducted using the Raisim\cite{raisim} simulator on a workstation with an AMD EPYC 9654 CPU and an NVIDIA RTX 4090 GPU. The training utilized 300 parallel environments for 80,000 iterations, taking approximately 1.7 seconds per iteration. In total, the entire training process required about 40 hours. 

For deployment, the robot was powered by an NVIDIA Jetson AGX Orin. The high-level policy was executed at 50~Hz and the low-level policy at 100~Hz, matching the simulation settings. To obtain exteroceptive information about the object and the table pose, we employed a Vicon Vero v2.2 motion-capture system.

\begin{figure}[!t]
\vspace{0.2cm}
\centering
\includegraphics[width=\columnwidth,trim={0cm 0cm 0cm 0cm}, clip]{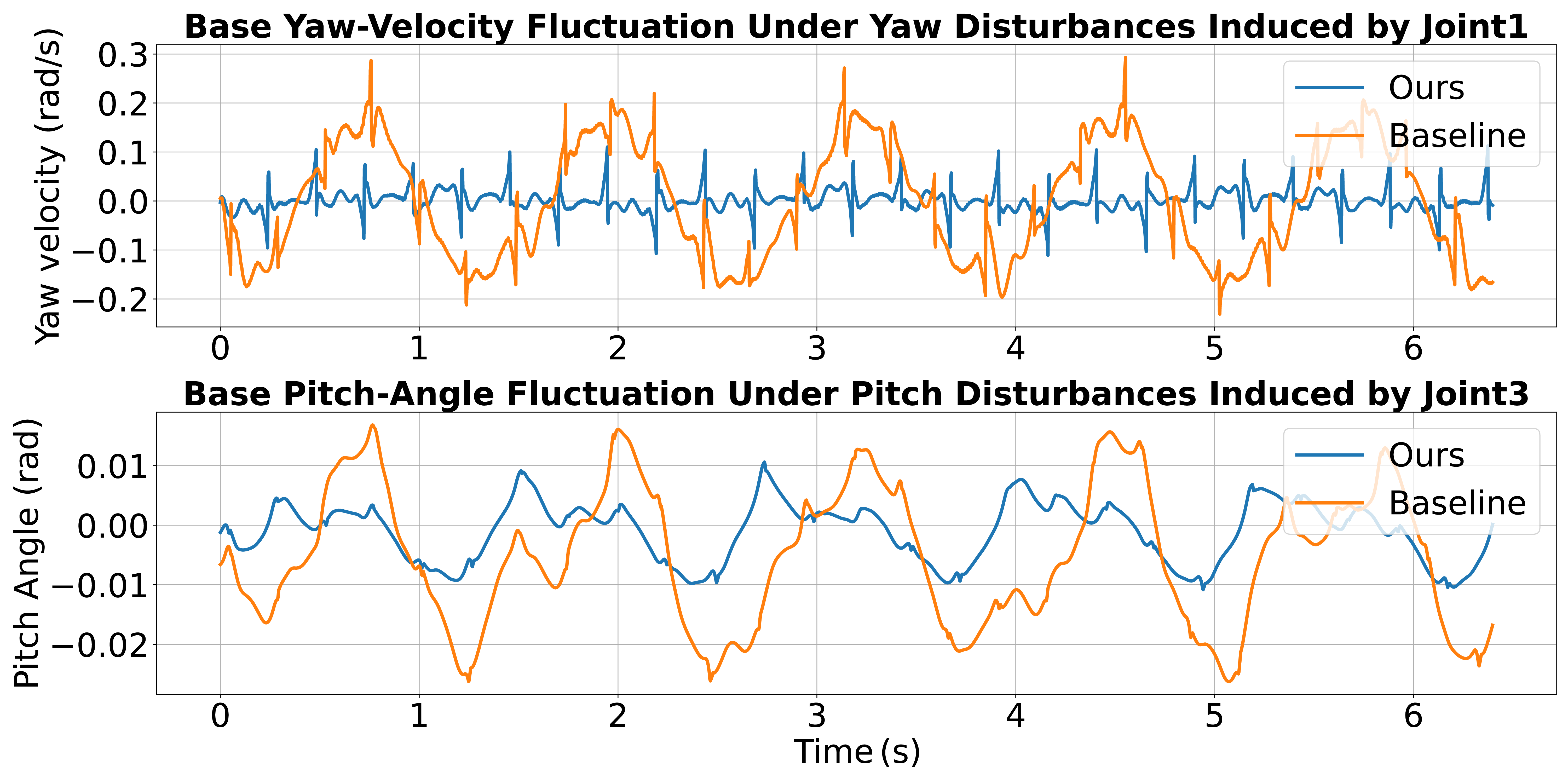}
\vspace{-0.8cm}
\caption[Disturbance rejection performance of the low-level controller]{
\textbf{Disturbance rejection performance of the low-level controller:}
\textit{Top}: Yaw-velocity fluctuation under yaw disturbances induced by periodic motion of Joint~1 (amplitude $1\,\mathrm{rad}$, frequency $0.8\,\mathrm{Hz}$). 
Our controller achieves a yaw-velocity RMSE of $0.026\,\mathrm{rad/s}$, compared to $0.117\,\mathrm{rad/s}$ for the baseline.
\textit{Bottom}: Pitch-angle fluctuation under pitch disturbances induced by periodic motion of Joint~3 (amplitude $1\,\mathrm{rad}$, frequency $0.8\,\mathrm{Hz}$). 
Our controller achieves a pitch-angle RMSE of $0.005\,\mathrm{rad}$, compared to $0.013\,\mathrm{rad}$ for the baseline.
}
\label{fig_low_distrubance}
\vspace{-0.5cm}
\end{figure}

\subsection{Evaluation of low-level controller(Simulation)}

The robustness of the low-level controller was evaluated by applying sinusoidal arm motions to individual joints under a forward-velocity command of $2~\mathrm{m/s}$. Specifically, we evaluated yaw robustness by applying periodic motion to Joint 1, and pitch robustness by applying periodic motion to Joint 3. We compared controllers trained with our joint-space random motion generator against a baseline that uses end-effector linear interpolation \cite{liu2024visual} followed by inverse kinematics.

Fig.~\ref{fig_low_distrubance} shows that our controller achieves substantially higher robustness for both yaw and pitch disturbance scenarios. The baseline, which relies on linearly interpolated end-effector trajectories, exhibits a narrower range of motion patterns and is constrained by the characteristics of the IK solver, which may limit its generalization under arm-induced disturbances.

\vspace{-0.3cm}

\begin{table}[!t]
\vspace{0.2cm}
\centering
\caption{Task Performance Comparison Across Baselines}
\label{table:performance}
{\tabcolsep=4pt\def\arraystretch{1.0}
\begin{tabular}{lcc}
\toprule
  \textbf{Architectures} & Success rate $[\%]~(\uparrow)$ & Completion Time $[s]~(\downarrow)$\tabularnewline 
\midrule
  Ours   & \textbf{86.05$\pm$0.43} & \textbf{2.071$\pm$0.004}\tabularnewline
  Latent adaptation & 85.82$\pm$0.37 & 2.098$\pm$0.005 \tabularnewline
  w/o estimation & 81.88$\pm$0.28& 2.082$\pm$0.008 \tabularnewline
\midrule 
Segmented Policy  & 78.40 $\pm$ 0.32 & 2.572 $\pm$ 0.014 \tabularnewline
\bottomrule
\end{tabular}}
\vspace{-0.3cm}
\end{table}

\begin{table}[!t]
\vspace{0.2cm}
\centering
\caption{Object Mass-wise Success Rate Comparison}
\label{table:masswise3}
{\tabcolsep=5pt\def\arraystretch{0.8} 
\begin{tabular}{lcccc}
\toprule
\multirow{2}{*}{\textbf{Object Mass} $[kg]$} & \multicolumn{2}{c}{\textbf{Success Rate} $[\%]$} & \multirow{2}{*}{\textbf{Gap} $[\%]$} \\
\cmidrule(lr){2-3} 
 & Ours & w/o Est. & \\
\midrule
$0.1 \sim 0.8$ & \textbf{84.36} & 80.80 & 3.56 \\
$0.8 \sim 1.5$ & \textbf{86.38} & 84.04 & 2.34 \\
$1.5 \sim 2.3$ & \textbf{84.98} & 81.66 & 3.32 \\
\rowcolor{gray!20} 
$2.3 \sim 3.0$ & \textbf{78.72} & 74.52 & 4.20 \\ 
\bottomrule
\end{tabular}}
\vspace{-0.4cm}
\end{table}

\subsection{Evaluation of high-level controller (Simulation)}
\subsubsection{Baseline Comparisons}

To assess the effectiveness of the proposed high-level controller in simulation, we conduct large-scale quantitative evaluations under diverse pick-and-place scenarios. We compare the proposed controller against three baseline variants using 5 random seeds. For each seed, we evaluate every method over 10,000 randomized episodes sampled from the training distribution, and report the mean and standard deviation across seeds. The baselines are defined as follows:

\begin{figure}[!t]
\vspace{0.1cm}
\centering
\includegraphics[width=\columnwidth,trim={32cm 10cm 36cm 0cm}, clip]{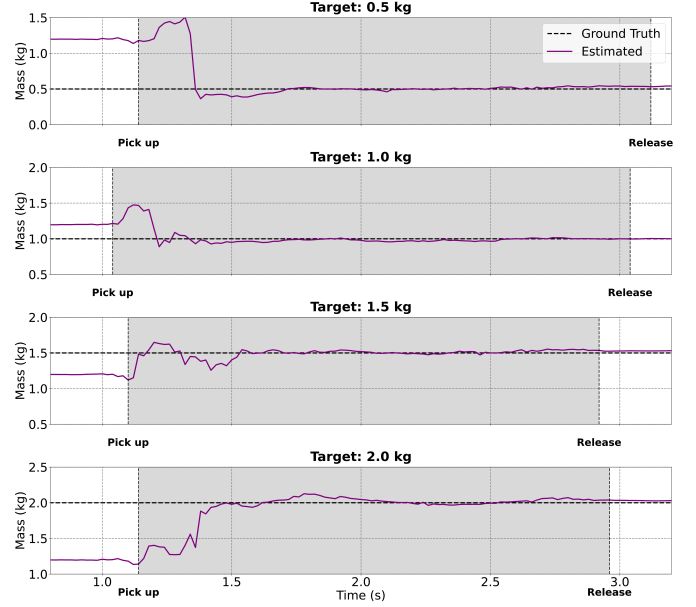}
\vspace{-0.6cm}
\caption[mass]{
\textbf{Real-time mass estimates during four pick–and–place episodes.}
The estimator begins each episode with an initial guess near the training-distribution mean of approximately 1.3~kg. After pick-up, the estimate shows a brief transient response before settling near the true object mass as the robot lifts and transports the object. The steady-state estimates at the release moment are [0.5319, 0.9973, 1.5350, 2.0385]~kg for the target masses of 0.5, 1.0, 1.5, and 2.0~kg, respectively.
}
\vspace{-0.65cm}
\label{fig:estimation}
\end{figure}

\begin{figure}[!t]
\vspace{0.2cm}
\centering
\includegraphics[width=\columnwidth,trim={0cm 0cm 0cm 0cm}, clip]{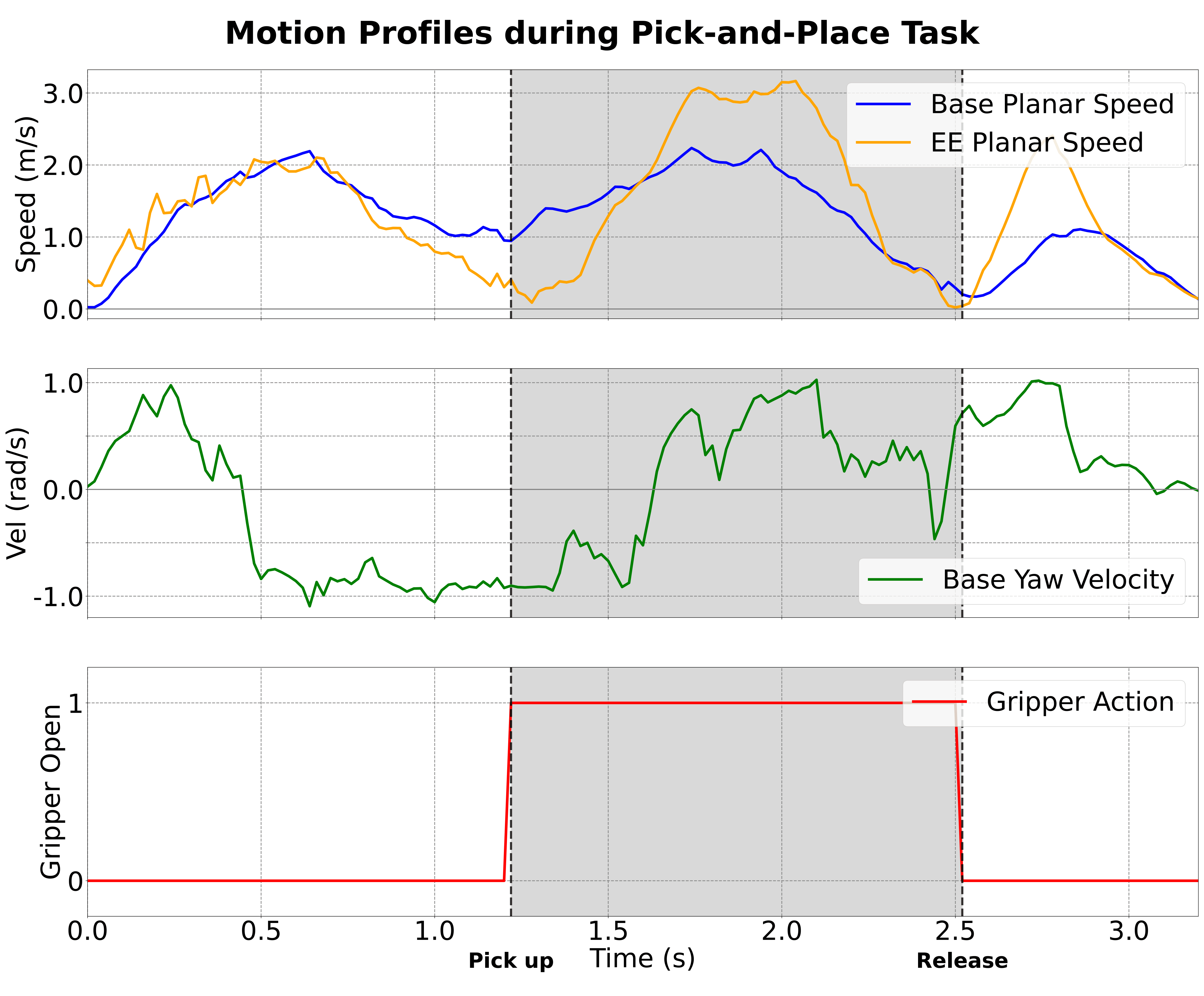}
\vspace{-0.6cm}
\caption[Base]{
\textbf{Base and end-effector motion profile in a single episode:} At the moment the gripper closes (rising edge of the gripper action, in red line), 
the quadruped base maintains a linear speed of approximately $1\,\text{m/s}$ 
(in blue line) and an angular velocity of about $-1\,\text{rad/s}$ 
(in green line), while the end-effector speed has already 
reduced to around $0.2\,\text{m/s}$ (in yellow line).
}
\vspace{-0.6cm}
\label{fig:timeline}
\end{figure}

\begin{itemize}

    \item \textbf{Latent Adaptation:}
    The LSTM-based mass estimator in our high-level controller is replaced with an encoder that performs adaptation in a latent space, following the latent representation approaches in \cite{fu2022deep, jeon2023whole}.
    
    \item \textbf{w/o Estimation:}
    A non-adaptive policy trained solely with domain randomization, where no LSTM estimator is used and the policy relies only on the current-step state.

   \item \textbf{Segmented Policy:} Inspired by the modular skill coordination philosophy of ASC \cite{yokoyama2024asc}, this baseline sequentially executes separately trained approach-and-grasp and transport-and-place policies
\end{itemize}

For evaluation, a successful episode requires placing the object upright within 5 cm of the place table center and retreating the end-effector by $\geq$10 cm without disturbing the object, all within a 10 s horizon.

Table~\ref{table:performance} summarizes the task performance. Compared to our unified controller, the segmented policy suffers a significant drop in success rate and execution speed. Without value propagation across task phases, the segmented grasping policy behaves myopically, optimizing only for immediate grasp completion while ignoring whether the resulting pose and object position are suitable for placement. This short-sightedness leads to quasi-static execution and increased collision rates when tables are placed closely.

Among the unified policies, the latent-adaptation baseline achieves performance comparable to ours; however, it relies on two encoders, which is less efficient for estimating low-dimensional variables such as mass and contact cues. In contrast, our explicit estimation module offers comparable performance with an 11.32\% reduction in training time per iteration, providing a more efficient solution with lower training cost. Furthermore, our controller outperforms the w/o-estimation baseline by about 4\%, with the gap widening as object mass deviates from the mean (Table~\ref{table:masswise3}). This effect is most pronounced in the shaded out-of-distribution mass range, where heavier objects require a larger lifting margin. Guided by real-time mass identification (Fig.~\ref{fig:estimation}), our policy proactively adjusts the lift height, whereas the non-adaptive baseline reacts only after insufficient motion is observed.

\subsubsection{Dynamic Base Motion During Grasping}

To further examine the motion behavior of the learned policy, we analyze the base movement during a representative dynamic case. 
Figure~\ref{fig:timeline} shows the linear and angular velocities of the quadruped base around the grasping moment.  As the end-effector velocity smoothly converges to zero at grasp, the base continues moving with substantial linear and angular motion—reaching up to $1\,\mathrm{m/s}$ planar speed and $-1\,\mathrm{rad/s}$ yaw rate in the fastest pick-and-place cases—allowing the robot to sustain locomotion without pausing. This illustrates the controller’s ability to coordinate dynamic whole-body motion:  the base provides continuous momentum for efficient task execution, while the arm executes the precise, low-velocity control required for reliable grasping.
\vspace{-0.3cm}

\begin{table*}[!t]
\centering
\vspace{0.2cm}
\caption{Task Performance Across Six Evaluation Scenarios: Real-World Results and Corresponding Simulation Results}
\label{table:real_world_2}
{\tabcolsep=8pt\def\arraystretch{0.8}
\begin{tabular}{l l c c c c c c c}
\toprule
\textbf{Scenarios} &  & \multicolumn{2}{c}{Nominal} & \multicolumn{5}{c}{Task Variations (Ours)} \\
\cmidrule(lr){3-4} \cmidrule(lr){5-9}
& & Ours & VBC \cite{liu2024visual}& Heavy Object & Light Object & Square Object & Large Size & Large Height Gap \\
\midrule
Success & Sim & 9/10 & 8/10 & 8/10 & 9/10 & 10/10 & 7/10 & 7/10 \\
Rate & Real & 9/10 & 3/10 & 8/10 & 8/10 & 6/10 & 7/10 & 6/10 \\
\midrule
Completion & Sim & 3.92$\pm$0.40 & 4.12$\pm$0.62  & 3.99$\pm$0.36 & 3.88$\pm$0.41 & 3.91$\pm$0.36 & 3.88$\pm$0.37 & 5.26$\pm$1.04 \\
Time $[s]$ & Real & 4.07$\pm$0.66 & 4.76$\pm$0.86 & 3.74$\pm$0.72 & 4.08$\pm$0.74 & 3.74$\pm$0.97 & 3.54$\pm$0.61 & 5.20$\pm$1.05 \\
\bottomrule
\vspace{-0.7cm}
\end{tabular}}
\end{table*}

\subsection{Evaluation of High-Level Controller (Real-World)}

For real-world experiments, reproducing the full range of randomized training conditions in every trial was infeasible. Instead, we defined a nominal pick-and-place setup—specified by object shape, size, mass, and table-height difference—and varied one factor at a time to construct six evaluation scenarios. For each scenario, we conducted ten trials and measured both success rate and total task completion time. The six evaluation scenarios are as follows:
\begin{itemize}
    \item \textbf{Nominal}: $\varnothing 6 \times 10~\mathrm{cm}$ cylinder, 0.83~kg mass, table heights 0.6~m $\leftrightarrow$ 0.9~m
    \item \textbf{Heavy Object}: 1.3~kg object mass
    \item \textbf{Light Object}: 0.47~kg object mass
    \item \textbf{Square Object}: $5\times5\times10~\mathrm{cm}$ square prism
    \item \textbf{Large Size}: $\varnothing 7 \times 10~\mathrm{cm}$ cylinder
    \item \textbf{Large Height Gap}: table heights 0~m $\leftrightarrow$ 1.1~m
\end{itemize}

In all scenarios, the pick and place tables were positioned 1~m from the global origin, corresponding to a separation of 2~m. The robot’s initial pose was randomized within a radius of 3~m from the origin in each trial. The test objects were built by inserting standard calibration weights inside 3D-printed housings, allowing controlled variations in size and mass. As our tracking system required attaching reflective markers to the object, we excluded objects too small to accommodate these markers. Although the system was trained with object masses up to 2.3~kg, the real-world experiments were limited to 1.3~kg because the gripper’s built-in safety mode restricted the available gripping torque, preventing the use of the maximum gripping torque specified in the manual.

Table~\ref{table:real_world_2} summarizes the real-world performance of the learned high-level controller across the six scenarios, together with the corresponding simulation results under the same conditions.
Overall, the robot achieved a 73.3\% success rate over 60 trials without any additional training or fine-tuning after deployment from simulation, demonstrating strong zero-shot transfer.
The controller also generalized well to mass variation: both the light (0.47~kg) and heavy (1.3~kg) objects achieved an 80\% success rate. The most challenging condition was the \textit{Large Height Gap}, where the robot lifted an object from the ground and placed it on a 1.1~m-high table. Despite requiring substantial whole-body coordination—such as base pitching and arm extension (Fig.~\ref{fig: intro})—the controller still achieved a 60\% success rate, demonstrating its ability to handle extreme kinematic and balance constraints. The average execution time across all real-world experiments was 4.06~s. In the \textit{Large Height Gap} scenario, the average completion time increased to about 5.2~s, reflecting the task difficulty of near-ground pick-up and high-table placement.

To further assess the importance of low-level robustness on real hardware, we evaluated a policy retrained with the baseline low-level controller from Section IV-A~\cite{liu2024visual} under the \textit{Nominal} condition. Although it achieved an 80\% success rate in simulation, its performance dropped to only 30\% on real hardware, revealing a substantial sim-to-real gap. These failures were mainly caused by arm motions destabilizing the base, often leading the robot to lose balance and topple over.

\textit{Failure Modes: }Most failures occurred during grasping. Although the end-effector usually reached a feasible grasping pose, the object often slipped or was ejected from the gripper as the fingers closed. Once a stable grasp was established, the robot generally completed the subsequent transport, placement, and arm retraction successfully. An exception was the \textit{Square Object} scenario, where two of the four failures occurred because the object tipped over during placement.
These failures can be attributed to three main factors.
First, a contact-model mismatch contributes to the large sim-to-real gap in the \textit{Square Object} scenario (100\% in simulation vs. 60\% in the real world).  In simulation, the gripper is approximated as a cuboid, which under-represents the edge and corner contacts that are particularly important for grasping prismatic objects.
Second, the gripper operates with binary open-close position commands, limiting its ability to adjust grasp force to object shape or friction. Third, near-ground manipulation in the \textit{Large Height Gap} scenario remains difficult because this challenging workspace was intentionally chosen near the kinematic limits of the platform, where feasible grasping angles are restricted and dynamic whole-body stabilization becomes more sensitive.

\section{CONCLUSION}
This work presents a hierarchical reinforcement learning framework that enables dynamic, whole-body pick-and-place with a legged manipulator by combining a robust low-level locomotion controller, a task-level high-level policy, an online mass-estimation module, and a success-rate-driven curriculum. The proposed system achieves rapid and reliable manipulation, characterized by an average real-world execution time of 4.06\,s and an overall success rate of 73.3\% across diverse scenarios involving payloads up to 1.3\,kg and vertical height gaps of 1.1\,m. These results demonstrate that the proposed framework can effectively coordinate agile locomotion and precise arm control to ensure high-speed task execution.

Future work will focus on eliminating the reliance on external motion-capture systems by incorporating ego-centric vision and extending the policy to generalize across a wider range of object geometries and physical properties.


%

\ifCLASSOPTIONcaptionsoff
  \newpage
\fi



%

\bibliographystyle{IEEEtran}
\bibliography{ref}




\end{document}